\documentclass[twocolumn,headings=small,abstract=true,10pt]{scrartcl}

\usepackage[ngerman,english]{babel}
\usepackage[T1]{fontenc}
\usepackage[utf8]{inputenc}
\usepackage{mathptmx}
\usepackage[scaled]{helvet}
\usepackage[sort&compress,numbers]{natbib}
\newcommand{\urlprefix}{}
\usepackage{amsmath,amsfonts}
\usepackage{microtype}
\usepackage[twocolumn,left=23mm,right=23mm,top=17mm,bottom=34mm]{geometry}
\usepackage{overpic}

\setkomafont{sectioning}{\normalfont\bfseries}
\setkomafont{caption}{\small}
\setcapindent{1em}
\usepackage{tabu}

\usepackage{ntheorem}

\theoremstyle{plain}
\theoremseparator{.}
\newtheorem{theorem}{Theorem}
\newtheorem{proposition}[theorem]{Proposition}
\newtheorem{lemma}[theorem]{Lemma}

\theoremstyle{definition}
\theorembodyfont{\normalfont}

\theoremstyle{remark}
\theoremheaderfont{\itshape}
\theorembodyfont{\normalfont}

\usepackage{hyperref}
\hypersetup{
  pdfauthor={Zijia Li, Josef Schicho, Hans-Peter Schröcker},  pdftitle={7R Darboux Linkages by Factorization of Motion Polynomials},  pdfkeywords={7R linkage, Darboux motion, factorization},    hidelinks=true,}

\newcommand{\R}{{\mathbb{R}}}
\renewcommand{\H}{{\mathbb{H}}}
\renewcommand{\DH}{{\mathbb{DH}}}
\newcommand{\cj}[1]{\overline{#1}}
\newcommand{\eps}{\varepsilon}
\newcommand{\qi}{{\mathbf{i}}}
\newcommand{\qj}{{\mathbf{j}}}
\newcommand{\qk}{{\mathbf{k}}}
\newcommand{\SE}[1][3]{\mathrm{SE}(#1)}
\newcommand{\Norm}[1]{\Vert #1 \Vert}
\providecommand{\newblock}{\relax}

\title{\Large 7R Darboux Linkages by Factorization of Motion Polynomials}
\author{
  \normalsize
  \begin{tabu}{X[c,0.33]X[c,0.33]X[c,0.33]}
    Z.~Li\linebreak RICAM\linebreak Austrian Academy of Sciences\linebreak Linz, Austria &
    J.~Schicho\linebreak RISC\linebreak Johannes Kepler University\linebreak Linz, Austria &
    H.-P.~Schröcker\linebreak Unit Geometry and CAD\linebreak University of Innsbruck\linebreak Innsbruck, Austria
  \end{tabu}
}
\date{\normalsize\today}

\begin{document}

\twocolumn[\begin{@twocolumnfalse}

\maketitle
\begin{abstract}
  \par\noindent    In this paper, we construct two types of 7R closed single loop linkages
  by combining different factorizations of a general (non-vertical) Darboux motion. 
  These factorizations are obtained by extensions of a factorization algorithm 
  for a generic rational motion. The first type of 7R linkages has several 
  one-dimensional configuration components and one of them corresponds to the 
  Darboux motion. The other type is a 7R linkage with two degrees of freedom and 
  without one-dimensional component. The Darboux motion is a curve in an irreducible 
  two dimensional configuration component.
 \end{abstract}
 \par\bigskip
\end{@twocolumnfalse}]

\section{Introduction}
\label{sec:introduction}

In 1881, G.~Darboux determined all one-parametric spatial motions with only planar curves as point paths
 \cite{darboux81}. These properties determine a unique motion which is nowadays named after Darboux. 
 It is the composition of a planar elliptic motion with a suitably oscillating translation 
 perpendicular to that plane. If the elliptic motion degenerates to a rotation, the Darboux motion 
 is called \emph{vertical,} if this is not the case, we call the Darboux motion \emph{non-vertical} 
 or \emph{general.} In both cases, the generic point paths are ellipses. 
 The Darboux motion is a special Schoenfließ motion and we call the direction perpendicular to the planar elliptic motion component the \emph{axis direction.}
 These and other properties are well known in literature, see for example \cite{blaschke58} or 
 \cite[Chapter~9, \S3]{bottema90}.

In this article, we construct 7R linkages with one degree of freedom such that one 
 link during (one component of) the motion performs a general Darboux motion. 
 These results complement recently presented linkages that generate vertical Darboux 
 motions \cite{lee12}. The beginnings of this research can be found in 
 \cite{li14:_spatial_straight_line_linkages} where we investigated spatial 
 linkages with a straight line trajectory. The linkages we present in this article are novel. 
 Unlike the linkages of \cite{lee12}, we generate a general (non-vertical) Darboux motion 
 and our linkages are free of prismatic or cylindrical joints. One of our linkages has already be 
 found in \cite{li14:_spatial_straight_line_linkages}. Our description here is more 
 general and detailed.

Our linkage construction is based on the factorization of rational motions 
 which has been first introduced in \cite{hegedus13:_factorization2}. 
 In the dual quaternion model of rigid body displacements, the Darboux motion 
 is parameterized by a cubic motion polynomial. In general, a cubic motion can 
 be generated by 6R linkages \cite{hegedus14:_four_pose_synthesis,li13:_anglesymmetric,li14:_ck} 
 but specialties of the Darboux motion prevent application of the 
 factorization algorithm for generic motion polynomials. Therefore, 
 we resort to new factorization techniques for non-generic motion polynomials 
 \cite{gallet15,li15:_factorization,li15:_minimal_degree}, i.e., we
 multiply the motion polynomial with a real 
 polynomial which
 divides the primal part of the motion polynomial. The new motion polynomial can be factored in different ways which allows us to obtain new closed 7R linkages 
 by combining different factorizations. A unified algorithm for factorization of
 generic and non-generic motion polynomials
  is introduced in \cite{li15:_factorization}. Here, we just mention the basic idea of this algorithm: By factoring out linear motion polynomial factors (from the left and from the right) we create a generic motion polynomial that can be factored by the original algorithm of \cite{hegedus13:_factorization2}.

We continue this text with a quick introduction to the dual quaternion model 
 of $\SE$, the group of rigid body displacements, and to motion polynomials 
 and their factorization (Section~\ref{sec:preliminaries}). 
 In Section~\ref{sec:factorizations} we describe the Darboux motion in the dual quaternion model, 
 present different factorizations, and combine them to form 7R linkages.

\section{Preliminaries}
\label{sec:preliminaries}

The investigations in this article are based on the dual quaternion model of the 
 group $\SE$ of rigid body displacements. We assume familiarity with dual 
 quaternions and their relation to kinematics but we provide a quick introduction 
 of concepts that cannot be found in standard textbooks. For more information on 
 basics of dual quaternions and kinematics we refer to
\cite{mccarthy90:_introduction_theoretical_kinematics,      selig05:_geometric_fundamentals_robotics,      husty12:_kinematics_algebraic_geometry}. A short presentation of the topic 
      with close relations to this article can also be found in \cite{li15:_survey}.
    
\subsection{Dual quaternions}

A dual quaternion is an object of the shape
\begin{equation}
  \label{eq:dualquat}
  h = h_0 + h_1\qi + h_2\qj + h_3\qk + \eps(h_4 + h_5\qi + h_6\qj + h_7\qk)
\end{equation}
with real numbers $h_0,\ldots,h_7$. The associative but non-commutative 
 multiplication of dual quaternions is defined via the relations
\begin{gather*}
  \qi^2 = \qj^2 = \qk^2 = \qi\qj\qk = -1,
  \quad
  \eps^2 = 0,\\
  \qi\eps = \eps\qi,\quad
  \qj\eps = \eps\qj,\quad
  \qk\eps = \eps\qk.
\end{gather*}
For example
\begin{equation*}
  \begin{aligned}
    (1 - \eps\qi)(\qj + \eps\qk) &= \qj + \eps\qk - \eps\qi\qj - \eps^2\qi\qk \\
                                 &= \qj + \eps\qk - \eps\qk - 0 = \qj.
  \end{aligned}
\end{equation*}
Conjugate dual quaternion and dual quaternion norm are
\begin{gather*}
    \cj{h} = h_0 - h_1\qi - h_2\qj - h_3\qk + \eps(h_4 - h_5\qi - h_6\qj - h_7\qk),\\
    \Norm{h} = h\cj{h}.
\end{gather*}
The primal part of $h$ is $p = h_0 + h_1\qi + h_2\qj + h_3\qk$ and the
dual part is $d = h_4 + h_5\qi + h_6\qj + h_7\qk$. Primal and dual
part are (ordinary) quaternions. The set of quaternions is denoted by
$\H$, the set of dual quaternions by $\DH:=\mathbb{D}\otimes_\R\H$. Writing $h = p + \eps d$ 
 with $p, d \in \H$, we have $\Norm{h} = p\cj{p} + \eps(p\cj{q}+q\cj{p})$.
Note that both $p\cj{p}$ and $p\cj{q}+q\cj{p}$ are real numbers and the 
 norm itself is a \emph{dual number.}

If the dual quaternion $h = p + \eps d$ has 
 a real norm
($p\cj{q}+q\cj{p} = 0$), its action on a point $[x_0,x_1,x_2,x_3]$ of
real projective three-space is defined via
\begin{equation}
  \label{eq:action}
  x \mapsto y = \frac{px\cj{p}+p\cj{q}-q\cj{p}}{p\cj{p}}
\end{equation}
where $x = x_0 + x_1\qi + x_2\qj + x_3\qk$,
$y = y_0 + y_1\qi + y_2\qj + y_3\qk$ and the image point has projective 
 coordinates $[y_0,y_1,y_2,y_3]$. The map \eqref{eq:action} describes 
 a rigid body displacement and the composition of such displacements corresponds 
 to dual quaternion multiplication. Note that $h$ and $\lambda h$ with 
 $\lambda \in \R \setminus \{0\}$ describe the same map.

The dual quaternion $h$ as in \eqref{eq:dualquat} describes a rotation 
 or translation, if and only if $h_4 = 0$. Among these, translations 
 are characterized by $h_1 = h_2 = h_3 = 0$. We speak 
 accordingly of \emph{rotation} or \emph{translation quaternions.} 
 In the former case, the revolute axis has Plücker coordinates $[h_1,h_2,h_3,-h_5,-h_6,-h_7]$. 
 In particular, the axis direction is $(h_1,h_2,h_3)$.

\subsection{Motion polynomials}

Now we study polynomials
\begin{equation}
  \label{eq:polynomial}
  C = c_nt^n + \cdots + c_1t + c_0
\end{equation}
in the indeterminate $t$ and with dual quaternion coefficients 
 $c_n,\ldots,c_0 \in \DH$. The set of these quaternions is denoted by $\DH[t]$. 
 Multiplication of polynomials in $\DH[t]$ is defined by the convention that 
 indeterminate $t$ and coefficients commute. The conjugate polynomial $\cj{C}$ is 
 the polynomial with conjugate coefficients $\cj{c_n},\ldots,\cj{c_0}$.

A polynomial $C \in \DH[t]$ is called a \emph{motion polynomial,} if its 
 leading coefficient $c_n$ is invertible (has non-zero primal part) and the 
 so-called norm polynomial $C\cj{C}$ has real coefficients. In this case, 
 it parameterizes a rational motion via the map \eqref{eq:action}. Note that 
 $C$ and $QC$ parameterize the same rational motion if $Q$ is a real polynomial 
 without zeros. This observation will be important in the next section.

The linear polynomial $C = t - h$ with $h$ as in \eqref{eq:dualquat} is
a motion polynomial if and only if $h_4 = h_1h_5 + h_2h_6 + h_3h_7 = 0$. 
 Assuming $(h_1,h_2,h_3) \neq (0,0,0)$, the displacements described by $C$ for 
 varying $t$ are rotations about a fixed axis but with different revolute angle. 
 The axis' Plücker coordinates are $[h_1,h_2,h_3,-h_5,-h_6,-h_7]$ and $(h_1,h_2,h_3)$ 
 is a direction vector of the axis.

A generic motion polynomial $C$ of degree $n$ admits $n!$ factorizations of the shape
\begin{equation}
  \label{eq:factorization}
  C = (t - h_1) \cdots (t - h_n)
\end{equation}
with rotation quaternions $h_1,\ldots,h_n$. The precise statement and a 
 factorization algorithm can be found in \cite{hegedus13:_factorization2}. 
 Here is suffices to say that a Darboux motion is not generic and the general 
 theory of \cite{hegedus13:_factorization2} gives no information about existence 
 or non-existence of factorizations.

Equation~\eqref{eq:factorization} admits an important kinematic interpretation: 
 Each factorization corresponds to a one-parametric motion of an open chain of $n$ 
 revolute joints whose end-effector follows the motion parameterized by~$C$. The 
 revolute axes are determined by the rotation quaternions $h_1,\ldots,h_n$. Different 
 factorizations may be combined to form closed loop linkages with this coupler motion.

In the remainder of this text we need auxiliary results related to the
factorization of motion polynomials. Their proofs can be found in
\cite{hegedus13:_factorization2}:

\begin{lemma}[Polynomial division]
  \label{lem:division}
  If $C$ and $D$ are polynomials in $\DH[t]$ and the leading
  coefficient of $D$ is $1$, there exist polynomials $Q,R \in \DH[t]$
  such that $C = QD + R$ and $\deg R < \deg D$.
\end{lemma}
Note that the computation of quotient $Q$ and remainder $R$ is possible by  
 straightforward polynomial long division in a non-commutative setting.

\begin{lemma}[Zeros and right factors]
  \label{lem:zero}
  The dual quaternion $h$ is a zero of the polynomial $C \in \DH[t]$ if and only 
  if there exists a polynomial $Q \in \DH[t]$ such that $C = Q(t-h)$.
\end{lemma}

The statement of Lemma~\ref{lem:zero} is less trivial than the corresponding 
 statement for real polynomials. In particular, with $C$ as in \eqref{eq:polynomial}, 
 the value of $C(h)$ (and hence also the term ``zero of $C$'') is \emph{defined} as $c_nh^n + \cdots + c_1h + c_0$. It is not 
 admissible to simply plug $h$ in a factorized representation of the shape \eqref{eq:factorization}. 
 With $C=(t-k)(t-h)$ and $h,k \in \DH[t]$ we have, for example, $C = t^2 - (h + k)t + kh$ and thus
\begin{equation*}
  C(h) = -kh + kh = 0
  \quad\text{but}\quad
  C(k) = hk + kh \neq 0.
\end{equation*}
Lemma~\ref{lem:zero} asserts that the ``usual'' relation between zeros linear 
 factors holds at least for \emph{right factors.}

\section{Factorizations of the Darboux motion}
\label{sec:factorizations}

Following \cite[Equation~(3.4)]{bottema90}, the general Darboux motion is given 
 by the parametric equations
\begin{equation*}
  \begin{aligned}
    X &= x \cos\varphi - y \sin\varphi \\
    Y &= x \sin\varphi + y \cos\varphi + a \sin\varphi \\
    Z &= z + b \sin\varphi + c(1 - \cos\varphi).
  \end{aligned}
\end{equation*}
Here, $\varphi \in [-\pi,\pi)$ is the motion parameter, $a$, $b$, $c$ are real 
 constants and $x$, $y$, $z$ resp. $X$, $Y$, $Z$ are coordinates in the moving and 
 the fixed frame. For $a = 0$, the motion is a vertical Darboux motion. In this paper 
 we exclude this case and assume $a \neq 0$.

Using the conversion formulas between matrix and dual quaternion representation of 
 rigid body displacements (see for example \cite{husty12:_kinematics_algebraic_geometry}), 
 we find the polynomial parameterization
\begin{multline}
  C_0 = (\qk + c\eps) t^3+(1 - \eps(a\qi - c\qk + b)) t^2\\
      + (\qk - \eps(a\qj + b\qk))t + 1
\end{multline}
for the Darboux motion. Left-dividing this motion polynomial by the leading 
 coefficient $\qk + c\eps$ (this amounts to a coordinate change in the fixed frame), we arrive at
\begin{equation}\label{dbmotion1}
  C = t^3 - (\qk - \eps(a\qj - b\qk))t^2 + (1 - \eps(b + a\qi - c\qk))t - \qk +c\eps.
\end{equation}
Our further investigations will be based on this parametric representation of the Darboux motion.

Note that the primal part of $C$ is $(t^2+1)(t - \qk)$. The presence of a real 
 factor violates the assumptions of \cite[Theorem~1]{hegedus13:_factorization2} 
 (existence of factorizations) and prevents application of Algorithm~1 
 (computation of factorizations) of that paper. Nonetheless, we will see below 
 that $C$ admits 
 infinitely many factorizations with three linear motion polynomials. 
 Moreover, suitable polynomial multiples of $C$ admit factorizations as well and parameterize 
 the same Darboux motion. This we will use for our construction of 7R Darboux linkages.

In \cite{li14:_spatial_straight_line_linkages}, the authors gave two types of 
 factorizations for the motion $C$ in Equation~\eqref{dbmotion1}. The first type is with three linear 
  motion polynomials (FI), the second type (FII) is with five linear motion polynomials 
  whose product does not equal $C$ but $C$ times a real quadratic polynomial $P$. This 
  makes an algebraic difference but does not change the kinematics: Both $C$ and $PC$ 
  parameterize the same motion. Moreover, two successive linear factors in FII are identical. 
  Hence, it gives rise to an open 4R chain which can generate the Darboux motion. In combination with 
  the 3R chain obtained from FI we obtain a 7R linkage.

In this section, we recall the construction of \cite{li14:_spatial_straight_line_linkages} 
 and we give two further factorizations (FIII and FIV) of \eqref{dbmotion1}. Similar to FII, 
 these factorizations also use five linear motion polynomials but give only four axes because 
 two successive linear motion polynomials are same. Combining it with the already known 3R chain, 
 we obtain further 7R Darboux linkages.

\subsection{First factorization FI}

We want to find linear motion polynomials $Q_1$, $Q_2$, $Q_3$ such that 
 $C = Q_1Q_2Q_3$. By Lemma~\ref{lem:zero}, a necessary and sufficient condition 
 for $Q_3 = t - q_3$, $q_3 \in \DH$, to be a right factor of $C$ is $C(q_3) = 0$. 
 A straightforward calculation shows that $q_3$ is necessarily of the shape
\begin{equation*}
  q_3 = \qk + \eps(v\qi + w\qj), \quad v, w \in \R.
\end{equation*}
Next, we consider the quotient polynomial $Q = P + \eps D$, defined by $C = QQ_3$. 
 Its primal part is $P = t^2+1$.  Thus, the motion parameterized by $Q$ is a curvilinear 
 translation. Since $Q$ is quadratic and $P$ has no real zero, the trajectories are congruent 
 ellipses. This motion admits a factorization $Q = Q_1Q_2$ of the required shape only 
 if it is a \emph{circular translation} (see \cite{li15:_survey} for a detailed proof of this statement). This 
 puts
 a condition on $q_3$, namely
\begin{equation}
  \label{eq:circular-translation}
  v = -a^{-1}bc,
  \quad
  w = - (2a)^{-1}(a^2+b^2-c^2).
\end{equation}

If \eqref{eq:circular-translation} is satisfied, the remainder $Q$ is a circular 
 translation and can be factored in infinitely many ways as $Q = Q_1Q_2$ with linear 
 rotation polynomials $Q_1$, $Q_2$. Each pair of these factorizations correspond to a 
 parallelogram linkage that generates the circular translation. One example of a 
 factorization of $C$ is
\begin{equation}\label{FI}
  \begin{aligned}
    Q_1 & = t + E - \frac{bc}{a} \qi \eps  + \frac{a^2+c^2-b^2}{2a} \qj \eps - b \qk \eps , \\
    Q_2 & = t - E, \\
    Q_3 & = t - \qk +\frac{bc}{a}  \qi\eps + \frac{a^2+b^2-c^2}{2a}  \qj\eps
  \end{aligned}
\end{equation}
where
\begin{equation}
  \label{eq:E}
  E = \frac{2ac}{a^2+b^2+c^2} \qi + \frac{2ab}{a^2+b^2+c^2} \qj + \frac{a^2-b^2-c^2}{a^2+b^2+c^2} \qk.
\end{equation}
In particular, we have shown

\begin{proposition}
  The general Darboux motion parameterized by \eqref{dbmotion1} can be decomposed 
  into a rotation about an axis parallel to the Darboux motion's axis direction and a circular 
  translation in a plane orthogonal to the vector quaternion $D$ of \eqref{eq:E}.
\end{proposition}

Combining two of the infinitely many factorizations with three linear factors gives 
 a parallelogram linkage to whose coupler a dangling link (corresponding to the right 
 factor $Q_3$ which is the same for all factorizations) is attached. This 6R linkage 
 has two trivial degrees of freedom and is not particularly interesting.

\subsection{Second factorization FII}

Now we construct a further factorization FII of the general Darboux motion. We do, 
 however, alter the parameterization \eqref{dbmotion1} and multiply it with the quadratic 
 polynomial $P := t^2+1$. This changes the motion polynomial but not the parameterized 
 motion and gives additional freedom to find factorizations. In contrast to FI, we have more 
 than three linear factors.

We begin by setting the right factor to $Q_4 := t - \qk$ and then use polynomial division 
 (Lemma~\ref{lem:division}) to compute
\begin{equation*}
    C_1 = t^2 + \eps(a\qj - b\qk)t + 1 + c\eps\qk.
\end{equation*}
such that $C = C_1Q_4$. The polynomial $PC_1$ can be written as the product of four 
 linear motion polynomials. Again, there exist infinitely many factorizations. In order 
 to keep the number of joints small, we choose one with identical middle factors, i.e., 
 $PC_1=Q_7 Q_6^2 Q_5,$ where
\begin{equation*}
  \begin{aligned}
    Q_7 & = t + \qi + \frac{a+c}{2} \eps\qj - \frac{b}{2} \eps\qk, \\
    Q_6 & = t - \qi,                                               \\
    Q_5 & = t + \qi + \frac{a-c}{2} \eps\qj - \frac{b}{2} \eps\qk.
  \end{aligned}
\end{equation*}
The multiplicity of the middle factor $Q_6$ allows us to make a 3R chain for the 
 motion parameterized by $PC_1$. Together with $C = Q_1Q_2Q_3$ and $Q_4$, we can get 
 a 7R linkage. It can be seen that the axes of $Q_1$, $Q_2$ are parallel, as are the 
 axes of $Q_3,\ Q_4$ and $Q_5$, $Q_6$, $Q_7$.

For a concrete example, a Gröbner basis computation reveals that the configuration 
 set of this 7R linkage is one dimensional \cite{myweb7R}. Note that its configuration 
 curve consists of different components and only one corresponds to the Darboux motion 
 parameterized by~$C$. 
  We also want to mention that the construction of the second factorization FII 
 leaves many degrees of freedom. We just presented one factorization of this type.

\subsection{Third factorization FIII}

In order to find a third factorization FIII that is suitable for linkage construction, 
 we proceed in a similar fashion to case FII,
 multiplying the motion polynomial $C$ of 
 Equation~\eqref{dbmotion1} with $P = t^2+1$. But we do so \emph{prior to splitting 
 off a right factor.}  The polynomial $PC$ has infinitely many factorizations into 
 products of \emph{five} linear motion polynomials, one of them being of the shape 
 $CP=Q'_7 Q'_6Q'_5Q'^2_4.$

We start by finding a suitable factor $Q'_4$. Because $Q'_4\cj{Q'_4}$ is a factor of 
 $(CP)(\cj{CP}) = P^5$, we have the necessary constraint $Q'_4\cj{Q'_4} = P$ but 
 still have infinitely many choices. We take
\begin{equation*}
  Q'_4 = t - \qk - x  \qi\eps - y  \qj\eps
\end{equation*}
with $x, y \in \R$. After division of $PC$ by $(Q'_4)^2$ (Lemma~\ref{lem:division}), 
 we are left with the cubic motion polynomial $C_2$ defined by $CP = C_2Q'^2_4$, i.e.,
\begin{multline*}
  C_2 = t^3 + (\qk + \eps(a\qj - b\qk + 2x\qi + 2y\qj)) t^2 \\
      + (1 + \eps(a\qi + c\qk + 2y\qi - 2x\qj + b))t + \qk - c\eps.
\end{multline*}
This new cubic motion polynomial $C_2$ parameterizes again a Darboux motion but the 
 parameterization is not in the standard form \eqref{dbmotion1}. Nonetheless, there 
  exist again infinitely many factorizations similar to \eqref{FI}. For example, we 
  have $C_2=Q'_7 Q'_6Q'_5$ where
\begin{equation*}
  \begin{aligned}
    Q'_5 &= t + \qk + \Bigl(\frac{b^2x-abc-2bcy-c^2x}{T^2+4x^2}+x\Bigr)\qi\eps \\
             & \quad + \Bigl( \frac{ab^2-ac^2+2b^2y+4bcx-2c^2y}{2(T^2+4x^2)}+\frac{a}{2}+y \Bigr)\qj\eps,\\
  Q'_6 &= t - \frac{2(ac-2bx+2cy)}{b^2+c^2+T^2+4x^2}\qi + \frac{2(ab+2by+2cx)}{b^2+c^2+T^2+4x^2}\qj\\
           & \quad + \frac{T^2-b^2-c^2+4x^2}{b^2+c^2+T^2+4x^2}\qk,\\
  Q'_7 &=\cj{Q'_6}  \phantom{+} \frac{Sx+bcT+4x(ay+x^2+y^2)}
              {T^2+4x^2}\qi\eps \\
        & \quad +
        \frac{(S+4x^2)aT+4(ay(a(3y+a)+2y^2)-abcx)}
             {2a(T^2+4x^2)}\qj\eps \\
       & \quad - b\qk\eps,\\
  \end{aligned}
\end{equation*}
and we abbreviated $S = a^2-b^2+c^2$ and $T = a + 2y$.

Because $C = Q_1Q_2Q_3$ and $PC =C_2Q'^2_4 = Q'_7Q'_6Q'_5Q'^2_4$ parameterize 
 the same motion, we can combine these two factorizations to form a 7R linkage 
 where each rotation is defined by $Q'_7$, $Q'_6$, $Q'_5$, $Q'_4$, $Q_3$, $Q_2$, $Q_1$.  
 It can be seen that the axes of $Q_1$, $Q_2$ are parallel, as are the axes of $Q_3$, $Q'_4$, 
 $Q_5$ and $Q'_7$, $Q'_6$.

As revealed by a Gröbner basis computation, the configuration space of a concrete numeric 
 example of this 7R linkage is really a curve \cite{myweb7R}. Thus, we have indeed constructed 
 another 7R linkage whose coupler motion is a non-vertical Darboux motion. Note that the 
 configuration curve contains several components, not all of them rational. One component 
 corresponds to the rational curve parameterized by~$C$.

\begin{figure*}
                \centering
                \begin{overpic}[width=0.49\textwidth,trim=0 32 0 45,clip]{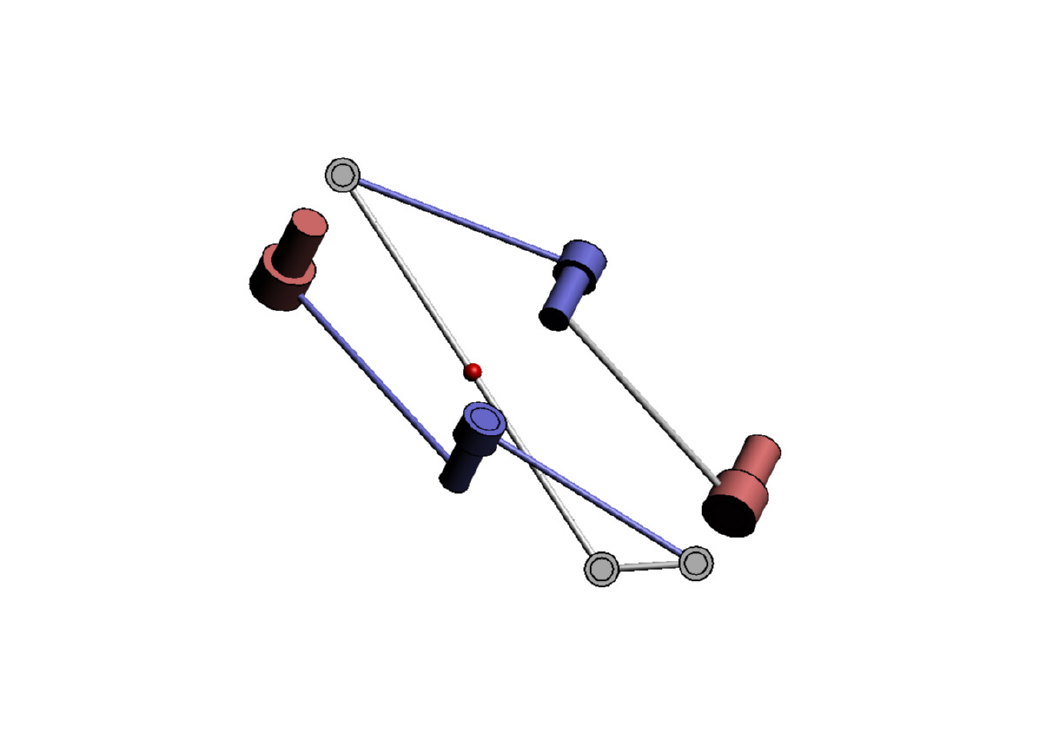}
                  \put(70,20){$Q_1$}
                  \put(57,37){$Q_2$}
                  \put(26,42){$Q_3$}
                  \put(68,4){$Q_4$}
                  \put(50,4){$Q_5$}
                  \put(38,11){$Q_6$}
                  \put(20,35){$Q_7$}
                \end{overpic}                \includegraphics[width=0.49\textwidth,trim=0 32 0 45,clip]{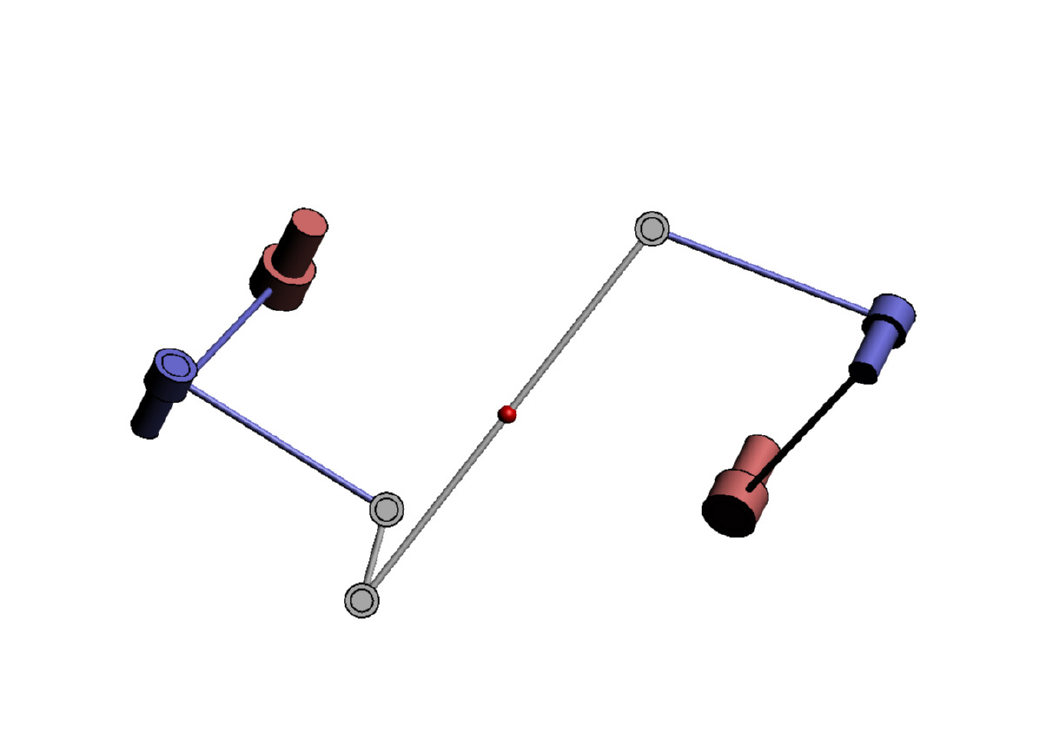}\\
                \includegraphics[width=0.49\textwidth,trim=0 28 0 28,clip]{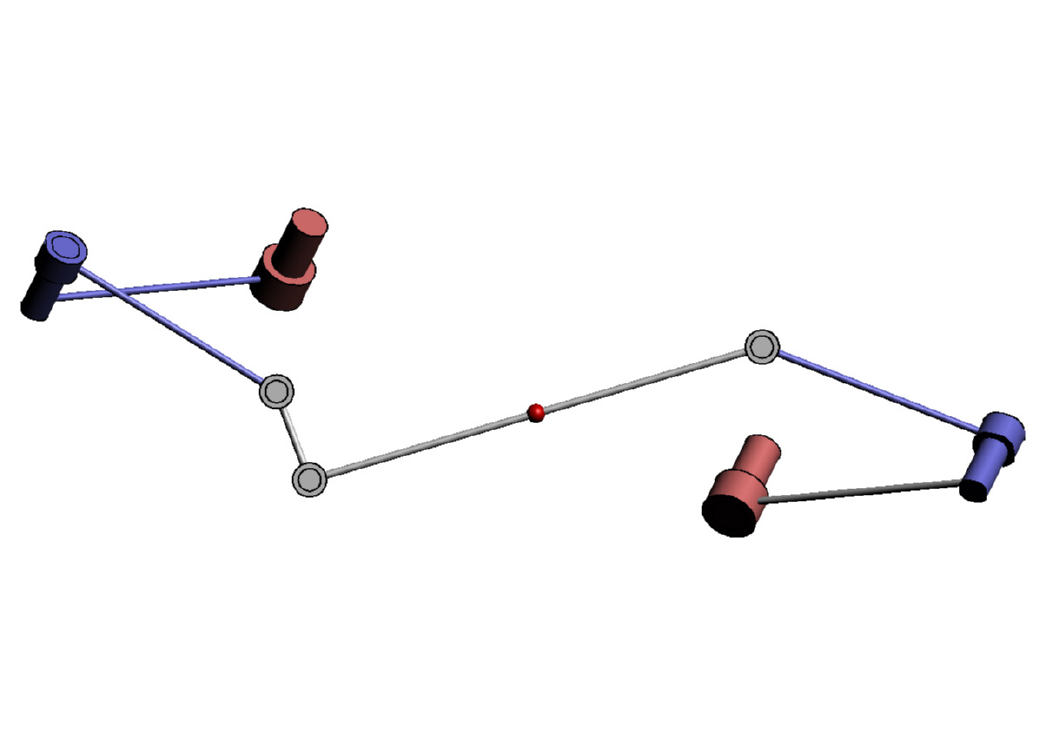}                \includegraphics[width=0.49\textwidth,trim=0 28 0 28,clip]{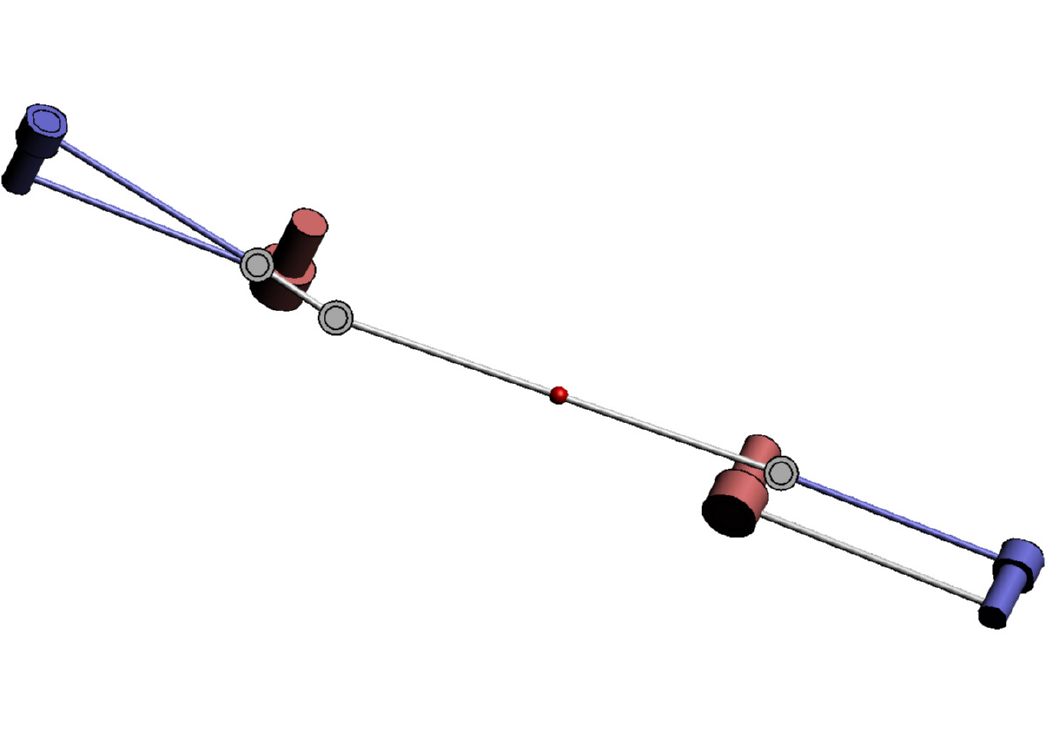}\\
                \includegraphics[width=0.49\textwidth,trim=0 0 0 0,clip]{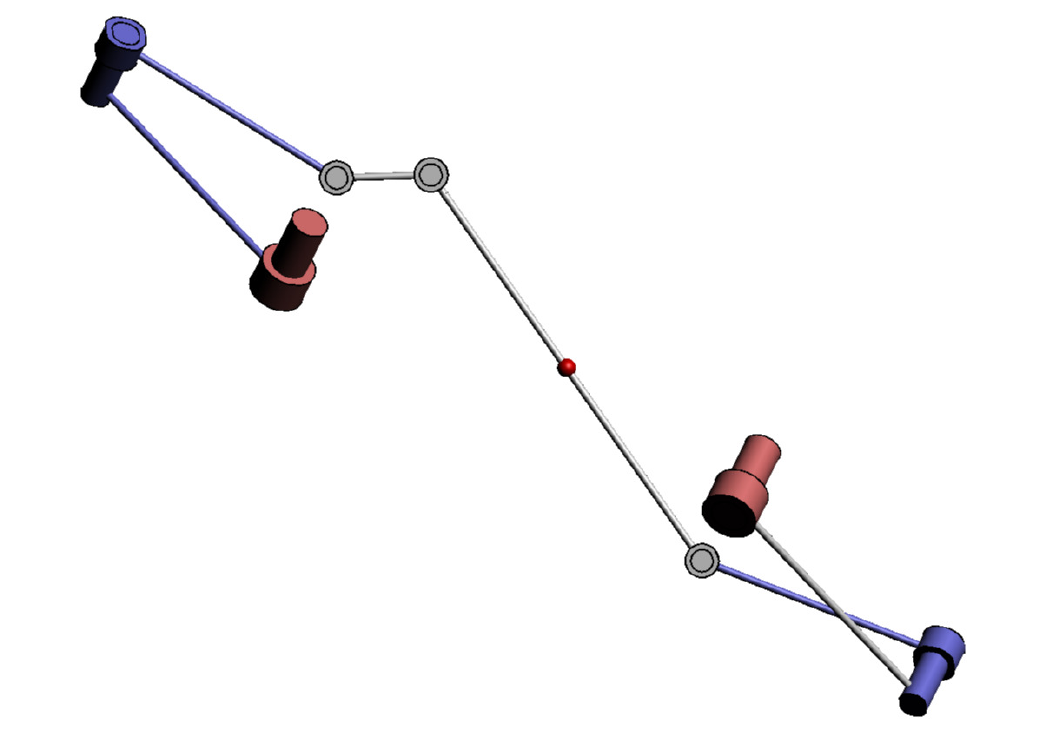}		\includegraphics[width=0.49\textwidth,trim=0 0 0 0,clip]{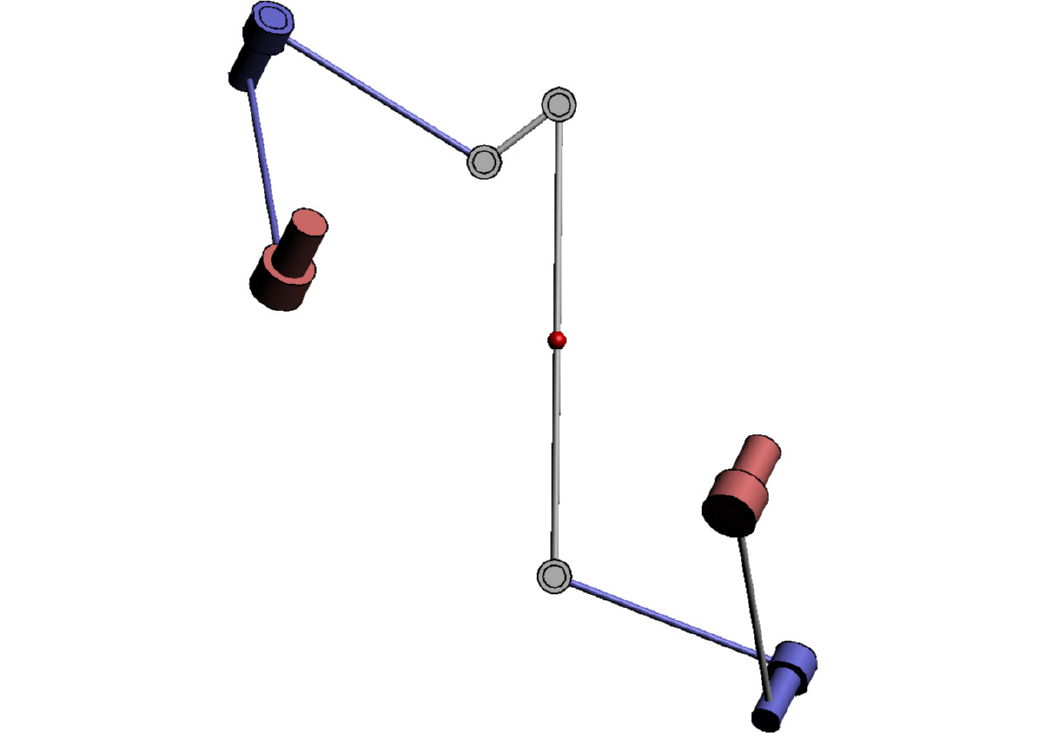}\\
                \includegraphics[width=0.49\textwidth,trim=0 17 0 18,clip]{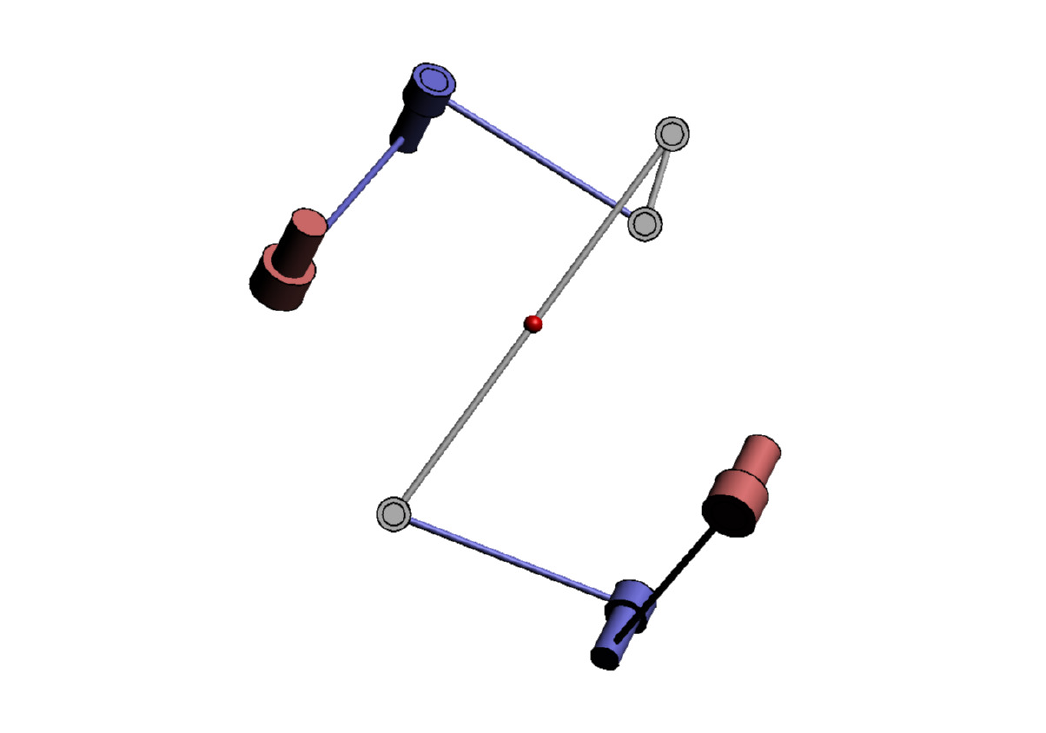}                \includegraphics[width=0.49\textwidth,trim=0 17 0 18,clip]{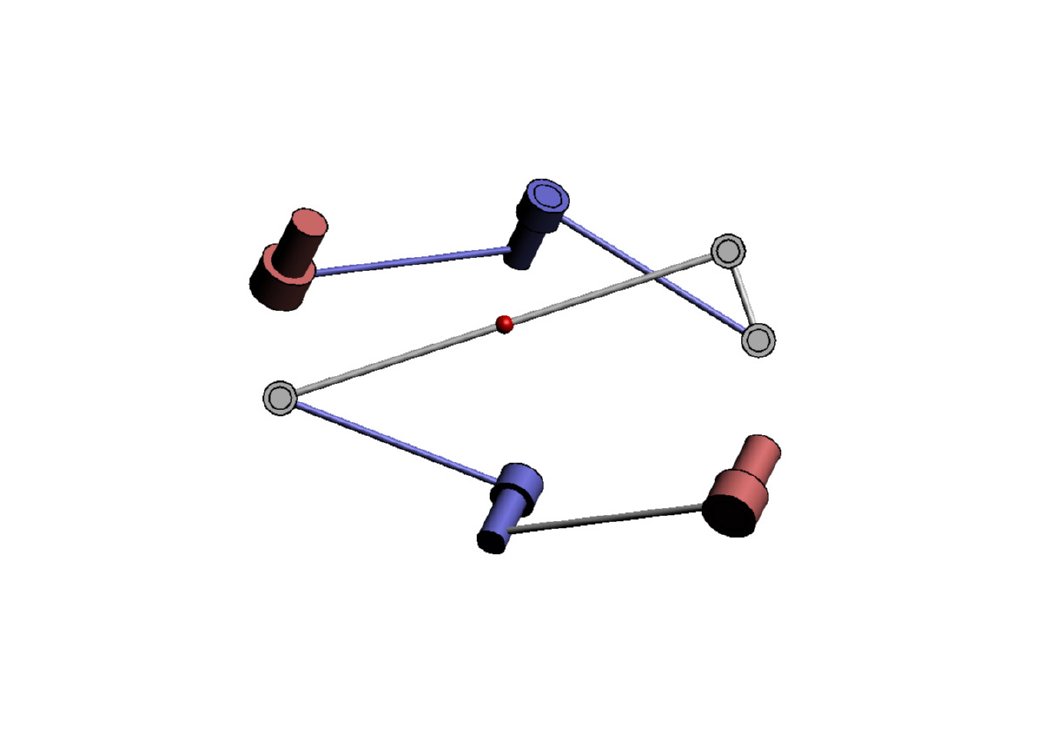}\\
                \includegraphics[width=0.49\textwidth,trim=0 40 0 43,clip]{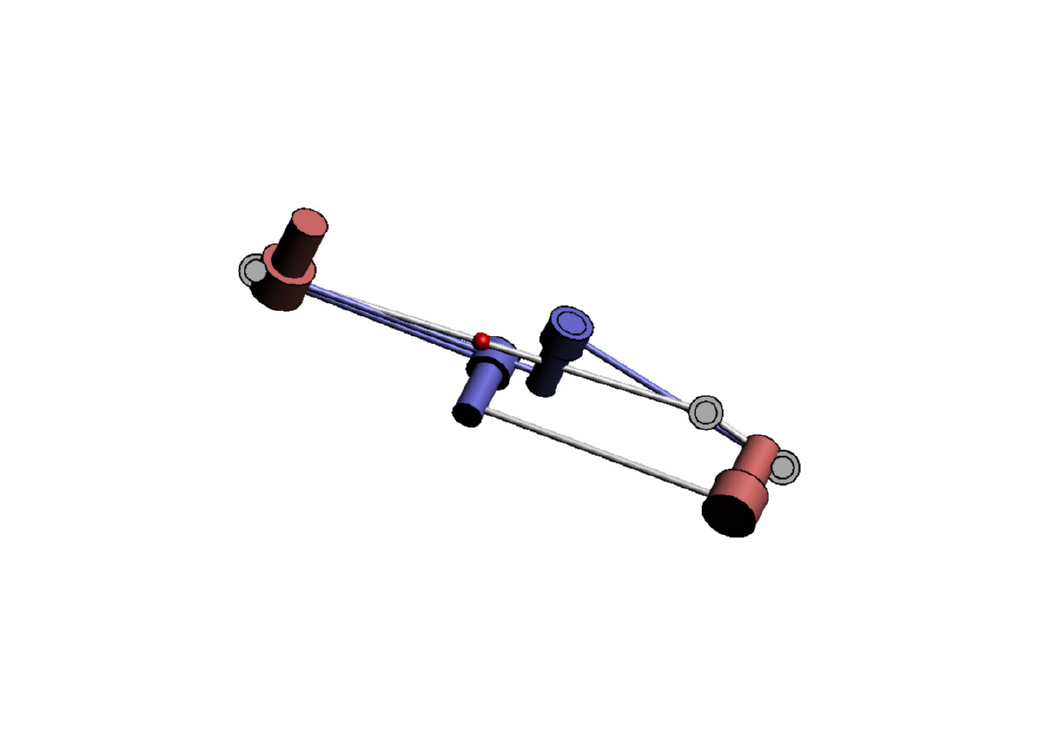}                \includegraphics[width=0.49\textwidth,trim=0 40 0 43,clip]{example4/example4+0}
  \caption{The first 7R linkage generates a non-vertical Darboux motion.}
  \label{fig:darboux1}
\end{figure*}

In Figure~\ref{fig:darboux1}, we present nine configurations (the first one and the last 
 one are from same configuration)  of this linkage in an orthographic projection parallel to~$\qk$.

\subsection{Fourth factorization FIV}

Finally, we present an exceptional case of the 7R linkage obtained
from the third factorization FIII. Among the infinitely many choices
for $a$, $b$, $c$, $x$, $y$ we choose the special case $a=1$, $b=2$,
$c=0$, $x=0$, $y=0$. We then have
\begin{equation*}
  \begin{aligned}
    Q''_1 & = t +   \frac{4}{5} \qj        - \frac{3}{5} \qk 
               -   \frac{3}{2}  \qj\eps -2 \qk\eps, \\
    Q''_2 & = t -   \frac{4}{5} \qj        + \frac{3}{5} \qk,       \\                                        
    Q''_3 & = t - \qk + \frac{5}{2}\qj\eps,\\
    Q''_4 & = t -   \qk, \\
    Q''_5 & = t + \qk + \frac{5}{2}\qj\eps,\\                                          
    Q''_6 & = t +   \frac{4}{5} \qj        - \frac{3}{5} \qk,  \\  
    Q''_7 & = t -   \frac{4}{5} \qj        + \frac{3}{5} \qk 
               -   \frac{3}{2}  \qj\eps -2 \qk\eps.
  \end{aligned}
\end{equation*}
We can combine this factorization to form a 7R linkage where each rotation is 
 defined by $Q''_7$, $Q''_6$, $Q''_5$, $Q''_4$, $Q''_3$, $Q''_2$, $Q''_1$.  It can be 
 seen that the axes of $Q''_1$, $Q''_2$, $Q''_7$, $Q''_6$ are parallel, as are the axes 
 of $Q''_3$, $Q_4''$, $Q''_5$.  Since four neighboring axes are parallel, this linkages 
 contains a planar 4-bar linkage. Furthermore, there are two pairs of parallel 3R sub chains 
 which give us two Sarrus linkages if we fix the remaining joint. Hence, this 7R linkage has 
 at least two degrees of freedom.

A Gröbner basis computation confirms that the configuration space is indeed two-dimensional 
 \cite{myweb7R}. Moreover, a decomposition of the configuration variety shows that the 
 linkage is not kinematotropic, which means that there are components of different dimension
 \cite{Galletti01:_kinematotropic}: 
 It consists of three irreducible 2-dimensional components. Two components are rational.
 We do not know whether the other two-dimensional 
 component which contains the rational curve parameterized by~$C$ is rational or not.

\begin{figure*}
                \centering
                \includegraphics[width=0.49\textwidth]{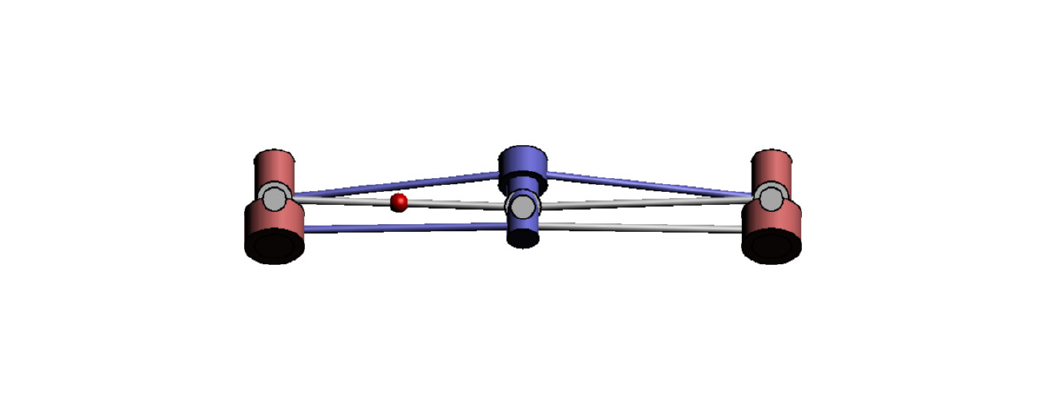}                \includegraphics[width=0.49\textwidth]{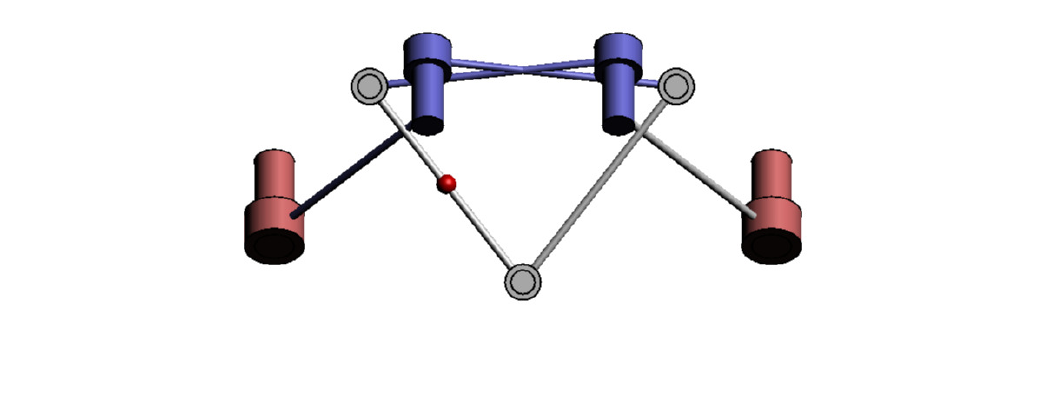}\\
                \includegraphics[width=0.49\textwidth]{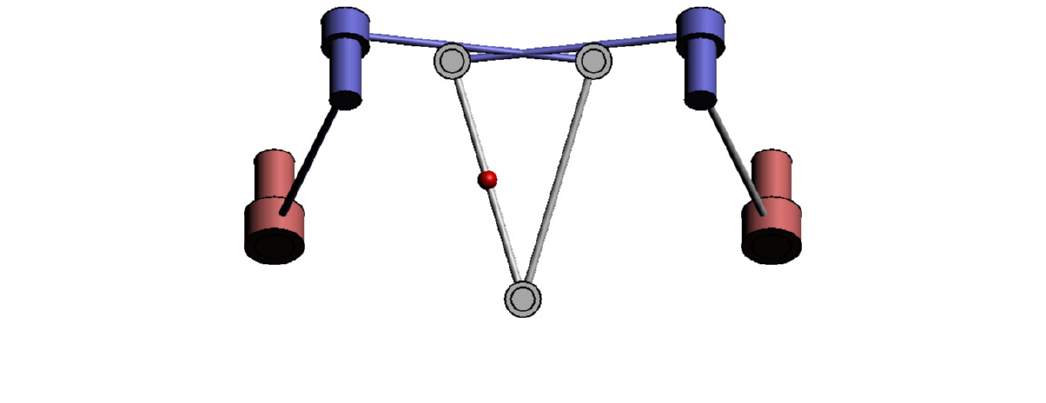}                \begin{overpic}[width=0.49\textwidth]{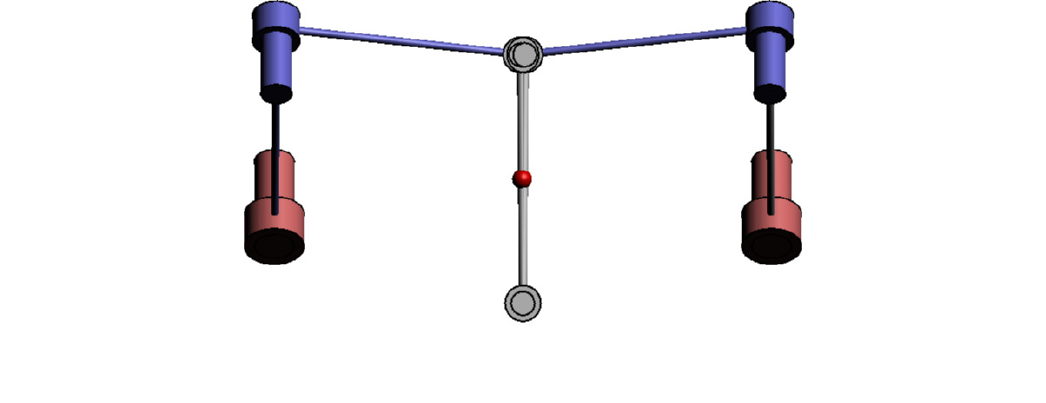}
                  \put(52,28){$Q''_3 = Q''_5$}
                \end{overpic}\\
                \includegraphics[width=0.49\textwidth]{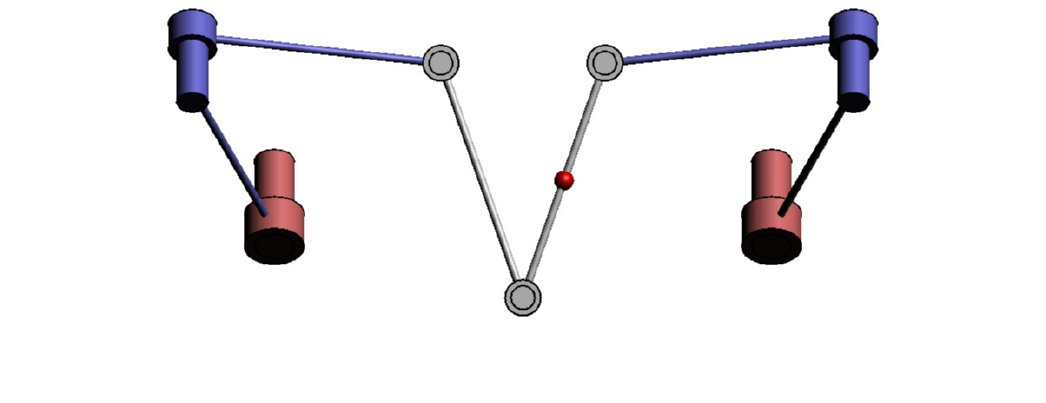}		\includegraphics[width=0.49\textwidth]{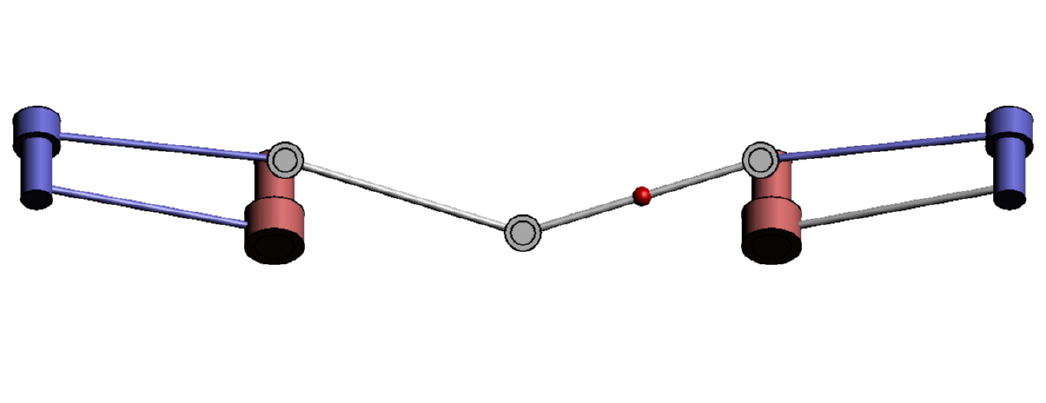}\\
                \includegraphics[width=0.49\textwidth]{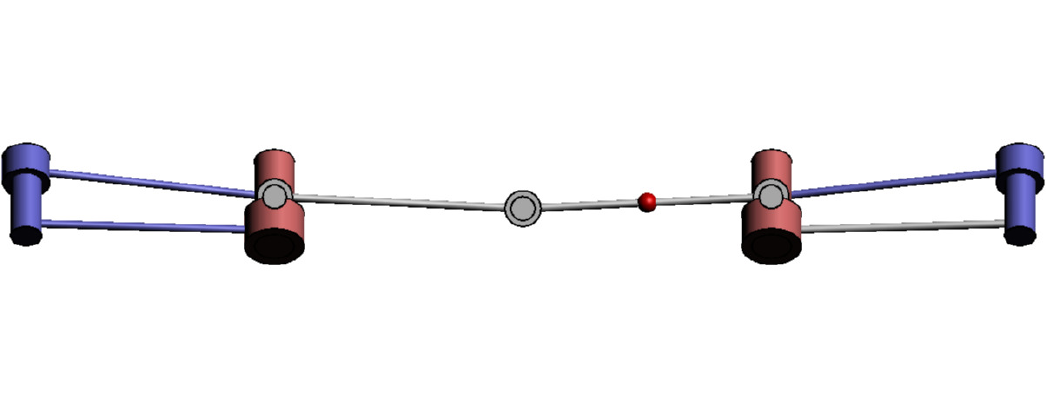}                \includegraphics[width=0.49\textwidth]{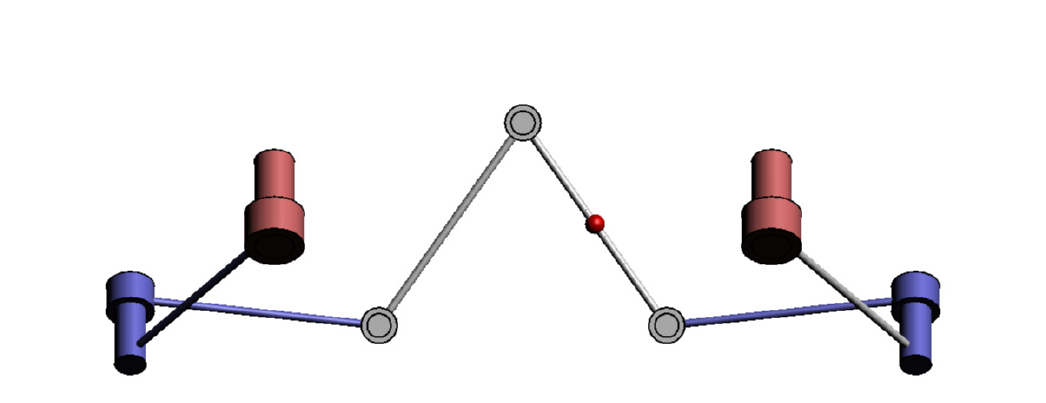}\\
                \begin{overpic}[width=0.49\textwidth]{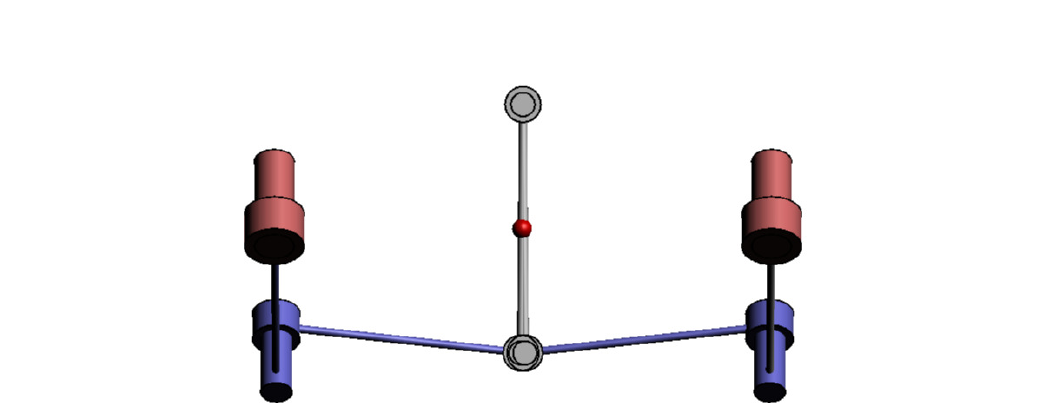}
                  \put(52,8){$Q''_3 = Q''_5$}
                \end{overpic}                \includegraphics[width=0.49\textwidth]{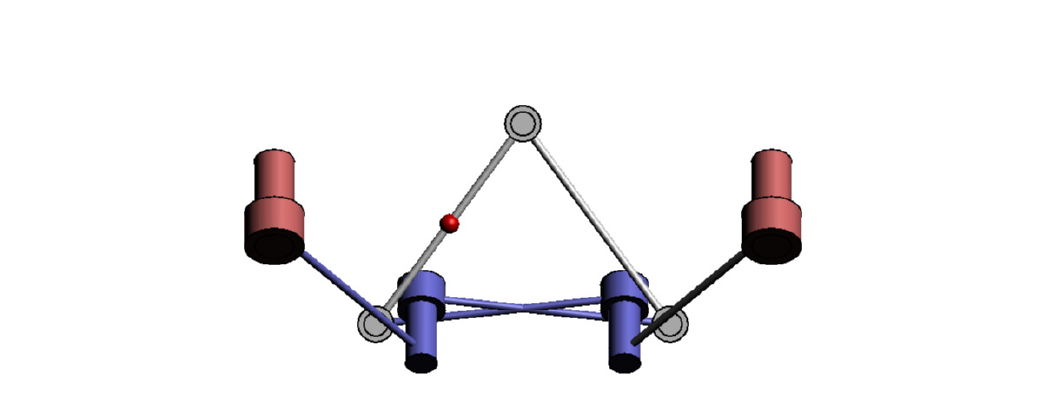}\\
                \includegraphics[width=0.49\textwidth]{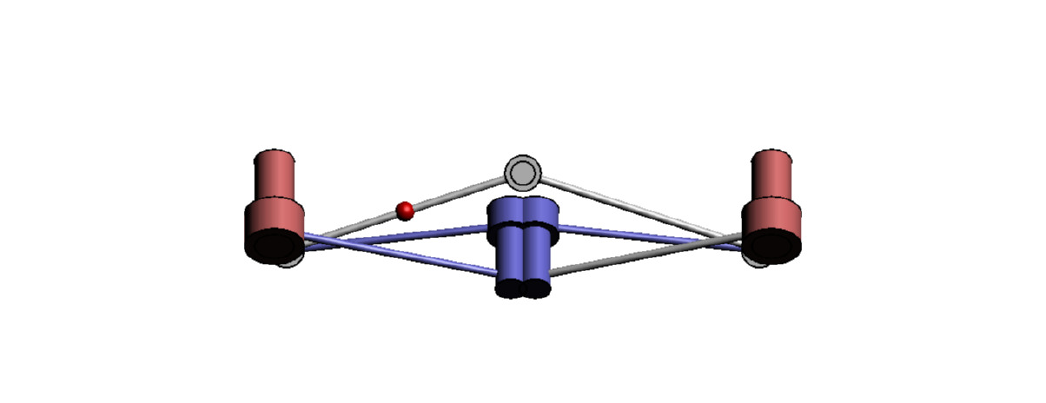}                \includegraphics[width=0.49\textwidth]{example5/example5+0}
  \caption{The second 7R linkage generates a non-vertical Darboux motion.}
  \label{fig:darboux3}
\end{figure*}

In Figure~\ref{fig:darboux3}, we present eleven configurations 
 (the first one and the last one are the same) of this linkage in 
 an orthographic projection parallel to the direction of ~$\qk$. 
 Because the axes of $Q''_1$, $Q''_2$, $Q''_7$, and $Q''_6$ are parallel, 
 the remaining axes are depicted as points throughout the motion. We can 
 observe the parallelity of axes and possible singularities (intersections). 
 The two configurations where the axes of $Q''_3$ and $Q''_5$ coincide give us 
 two rational configuration sets of dimension two: A planar 4R loop 
 (the coupler motion is a circular translation) and a single rotation 
 rational component. None of these two  contains the Darboux motion.

\section{Conclusions}
\label{sec:conclusions}

Using factorization of motion polynomials in non-generic cases, we constructed several examples of 
 new 7R Darboux linkages. The main idea is to find an irreducible real polynomial $P$ and 
  factor $PC$ instead of $C$. The obtained 7R linkages generate the general (non-vertical) 
   Darboux motion and exhibit some interesting specialities like parallel axes or 
   circular translations as relative motions between certain links. One can replace two R-joints by other 
   R-joints which generate
   same circular translations.

Verifying the validity of the presented factorizations and corresponding linkages is 
 trivial but, admittedly, we did not explain in detail how to actually find the 
 factorizations. The factorization theory for non-generic motion polynomials, 
 including theoretical results, algorithms and upper bounds for the number of factors, 
 is currently being worked out. Its presentation is left to future publications.

\section*{Acknowledgements}

This research was supported by the Austrian Science Fund (FWF):
P\;26607 (Algebraic Methods in Kinematics: Motion Factorisation and
Bond Theory).

\bibliographystyle{iftomm}

\end{document}